\documentclass[letterpaper, 10 pt, conference]{ieeeconf}  % Comment this line out if you need a4paper

\IEEEoverridecommandlockouts                              % This command is only needed if 
\usepackage{xcolor}
\newtheorem{lemma}{Lemma}
\newtheorem{theorem}{Theorem}
\newtheorem{prop}{Proposition}
\usepackage{graphicx}
\usepackage{hyperref}
\usepackage{amsmath,amssymb,amsfonts,booktabs,multirow,bm,balance,bbm,siunitx,hyperref,graphicx}
\usepackage[caption=false, font=footnotesize]{subfig}
\usepackage[colorinlistoftodos]{todonotes}
\usepackage{float}
\let\labelindent\relax
\usepackage{enumitem,kantlipsum}
\usepackage{afterpage}
\usepackage{setspace}

\usepackage{makecell}
\usepackage{cellspace}

\usepackage[skip=10pt,font=small]{caption}
\usepackage{balance}   % To balance the columns on the last page
\usepackage{xcolor}

\title{\LARGE \bf
Predictive Control for Chasing a Ground Vehicle using a UAV
}

\author{Jaeseung Byun*, Karan P. Jain*, Siddharth H. Nair*, Haoyun Xu*, and Jiaming Zha*
\thanks{The authors are with the Department of Mechanical Engineering, University of California, Berkeley. CA, 94720. \{jaeseungbyun, karanjain, siddharth\_nair, haoyun95, jiaming\_zha\}@berkeley.edu}
\thanks{*All authors contributed equally to this project.}
}

\begin{document}

\maketitle
\thispagestyle{empty}
\pagestyle{empty}

%%%%%%%%%%%%%%%%%%%%%%%%%%%%%%%%%%%%%%%%%%%%%%%%%%%%%%%%%%%%%%%%%%%%%%%%%%%%%%%%
\begin{abstract}

We propose a high-level planner for a multirotor to chase a ground vehicle, while simultaneously respecting various state and input constraints. Assuming a minimal kinematic model for the ground vehicle, we use data collected online to generate predictions for our planner within a model predictive control framework. Our solution is demonstrated, both via simulations and experiments on a stable quadcopter platform. 

\end{abstract}

%%%%%%%%%%%%%%%%%%%%%%%%%%%%%%%%%%%%%%%%%%%%%%%%%%%%%%%%%%%%%%%%%%%%%%%%%%%%%%%%
\section{Introduction}
%%%%
%Motivation
Multirotor UAVs are expected to be applied widely in the realm of urban air mobility\cite{silva2018vtol}. An interesting application in particular is the use of multirotors for chasing (or more formally, tracking) dynamic targets, owing to their agility and ubiquity. However, flying such vehicles in the city necessitates precise trajectory tracking of the vehicle and a framework that respects constraints-- those on the vehicle itself, as well as those imposed by the environment. Solutions to certain aspects of the problem have been proposed in the literature; \cite{hoang2017vision} and \cite{teuliere2011chasing} focus on  vision-based strategies to localize and track the target. More recently, \cite{archive2019} proposes a hybrid decision planner for docking UAVs on ground vehicles. We address the problem of trajectory and control design for UAVs to chase ground vehicles while adhering to constraints.
%%%%%

%%%
%%Lit Survey%%
In this paper, we present a modular solution to the problem that can be applied to any commercial UAV platform with minimal experimental tuning- a high-level planner that
\begin{enumerate}[label=\alph*)]
\item incorporates data from the past to predict the position of the ground vehicle and,
\item use the same data to generate waypoints that the quadcopter can follow without violating state and input constraints.
\end{enumerate}
 
For (a), the vehicle position is predicted by assuming a kinematic model with bounded states, and updating the bounds from past data using exponential moving average. These continually updated bounds are used to construct a set of predicted positions, of which, the Chebyshev center (\cite{borrelli2017predictive}) is used as the estimate of the vehicle's future position. In the context of (b), we use model predictive control (MPC), a popular strategy for synthesizing controls for systems with constraints, for generating the waypoints by solving a constrained finite time optimal control problem online \cite{borrelli2017predictive}. While the latter, in general, can be intractable for online computation owing to the often nonlinear dynamical behaviour of UAVs, we model our UAV in closed-loop with a low-level planner as a simple LTI system, thus making real-time, online computations feasible. We also present the control and communication architecture that enabled us to deploy our algorithm. 

%%Organisation of Paper%%
The rest of the paper is organized as follows. Section II formally states the problem along with the working assumptions before we present the theoretical framework for our solution in section III. Sections IV and V present simulations and experiments respectively that corroborate the effectiveness of our solution.

\section{Problem Formulation}
We work with a quadcopter equipped with a low-level controller that linearizes the dynamics around the hover condition, and is capable of driving the quadcopter to specified way-points. To chase the ground vehicle, we aim to generate way-points for the quadcopter that can be reasonably tracked using the low-level controller. Towards this goal, we list the blanket assumptions and models we plan to use for this work, and conclude this section with a formal statement of the problem.

%%%%%%%%%%%%%%%%%%%%%%%%%%%%%%

\subsection{Assumptions}\label{assump}
\begin{itemize}
    \item Accurate measurements of the states of the quadcopter are available.
    \item Accurate measurements of the position and velocity of the ground vehicle are available.
    \item Quadcopter is stable in hover with a low-level controller in loop.
    \item External disturbances are ignored owing to experiments being conducted in a spacious indoor testing arena.
    \item The velocity of the ground vehicle is bounded, and these bounds are known. 
    
\end{itemize}

%%%%%%%%%%%%%%%%%%%%%%%%%%%%%%

\subsection{Problem Statement}

Assuming that the conditions in \ref{assump} hold, design a control strategy for a quadcopter UAV to chase an evasive ground vehicle while adhering to specified state and input constraints.

%%%%%%%%%%%%%%%%%%%%%%%%%%%%%%
%%%%%%%%%%%%%%%%%%%%%%%%%%%%%%
%%%%%%%%%%%%%%%%%%%%%%%%%%%%%%

%\newpage
\section{Predictive Control for Trajectory Generation}

\noindent We adopt a hierarchical control strategy:
\begin{itemize}
    \item Off-board Control System: Use MPC to synthesize trajectories and send way-points to the quadcopter
    \item Onboard Control System: Drive the quadcopter towards the commanded way-points
\end{itemize}
The two parts are interconnected with radio/telemetry. In this section, we focus on the former and describe our approach to solve the problem.

%%%%%%%%%%%%%%%%%%%%%%%%%%%%%%

\subsection{Quadcopter Model}
We model the quadcopter with the on-board, inner control loop along the lines of \cite{bouffard2012board} and \cite{ccta}. The yaw angle is assumed fixed while the closed-loop roll and pitch dynamics are modelled as decoupled, second-order systems with input being a commanded angle (roll or pitch) and output being the actual angle (roll or pitch).
\begin{equation*}
G_{roll}(s)=\frac{a_r}{s^2+b_{r1}s+b_{r0}}
\end{equation*}    
\begin{equation*}
G_{pitch}(s)=\frac{a_p}{s^2+b_{p1}s+b_{p0}}    
\end{equation*}
The parameters $a_r,a_p,b_{r1},b_{r0},b_{p1},b_{p0}$ are obtained by system identification.\\
Translational motion along the x axis is assumed to be coupled only with pitch and the same along the y axis is assumed to be coupled only to roll, and are thus, decoupled from each other. Moreover, we assume that the attitude dynamics in closed loop are significantly faster than the translational (specifically altitude) dynamics. So the thrust at any instant balances the weight and is given by 
$$T=\frac{mg}{\cos\gamma}\quad \gamma={\theta, \phi}$$
So the lateral translational dynamics are given by
$$\ddot{x}=g\theta$$
$$\ddot{y}=-g\phi$$
Finally, the altitude dynamics are simply given by 
$$\ddot{z}=\frac{T_z}{m}-g$$
To consolidate, the quadcopter is described by the states and inputs $$\mathbf{X}=[ x\ \dot{x}\ \theta\ \dot{\theta}\ y\ \dot{y}\ \phi\ \dot{\phi}\ z\ \dot{z}]^T$$
$$\mathbf{U}=[\theta_{cmd}\ \phi_{cmd}\ T_z]^T$$
and the state space model in continuous time is expressed as 
$$\mathbf{\dot{X}}=A\mathbf{X}+B\mathbf{U}+G$$
where
\scriptsize
$$
A=
\begin{bmatrix}
0 & 1 & 0 & 0 & 0 & 0 & 0 & 0 & 0 & 0\\
0 & 0 & g & 0 & 0 & 0 & 0 & 0 & 0 & 0\\
0 & 0 & 0 & 1 & 0 & 0 & 0 & 0 & 0 & 0\\
0 & 0 & -b_{p0} & -b_{p1} & 0 & 0 & 0 & 0 & 0 & 0\\
0 & 0 & 0 & 0 & 0 & 1 & 0 & 0 & 0 & 0\\
0 & 0 & 0 & 0 & 0 & 0 & -g & 0 & 0 & 0\\
0 & 0 & 0 & 0 & 0 & 0 & 0 & 1 & 0 & 0\\
0 & 0 & 0 & 0 & 0 & 0 & -b_{r0} & -b_{r1} &0 & 0\\
0 & 0 & 0 & 0 & 0 & 0 & 0 & 0 & 0 & 1\\
0 & 0 & 0 & 0 & 0 & 0 & 0 & 0 & 0 & 0\\

\end{bmatrix}
$$

$$
B=
\begin{bmatrix}
0&0&0\\
0&0&0\\
0&0&0\\
-a_p&0&0\\
0&0&0\\
0&0&0\\
0&0&0\\
0&-a_r&0\\
0&0&0\\
0&0&\frac{1}{m}\\

\end{bmatrix}
$$
$$
G=[0\ 0\ 0\ 0\ 0\ 0\ 0\ 0\ 0\ -g]^T
$$

\normalsize
Assuming ZOH, we compute the exact discretization of these equations.
\begin{equation}\label{qmodel}
    \mathbf{X}_{k+1}=A_T\mathbf{X}_k+B_T\mathbf{U}_k +G_T
\end{equation}
where the matrices are defined below for discretization step $\Delta T$
\small
$$A_T=e^{A\Delta T}\quad B_T=(\int_0^{\Delta T}e^{A(\Delta T-\tau)}d\tau)B $$ $$G_T=(\int_0^{\Delta T}e^{A(\Delta T-\tau)}d\tau)G $$
\normalsize

%%%%%%%%%%%%%%%%%%%%%%%%%%%%%%

% \subsection*{System Identification Strategy}
% \par
% In order to experimentally identify the parameters $a_r,a_p,b_{r1},b_{r0},b_{p1},b_{p0}$, we observe that the transfer function that relates the acceleration in the x (or y) direction to the commanded pitch (or roll) is a second order system given by $gG_{pitch}(s)$ (or $g G_{roll}(s)$). Given that we have reliable measurements of acceleration of the quadcopter, we can identify the necessary parameters by analyzing the translational accleration responses to a unit step in the corresponding commanded angles.

%%%%%%%%%%%%%%%%%%%%%%%%%%%%%%

\subsection{Vehicle Model}
We assume a point mass kinematic model for the vehicle and we measure the position and velocity of the same. The states and inputs of the vehicle are given by $\mathbf{X}_v=[x_v\ y_v]^T$ and $\mathbf{U}_v=[v_x\ v_y]^T$ dynamics are written as
\begin{equation}
    \mathbf{\dot{X}}_v=\mathbf{U}_v
\end{equation}

In discrete time,
\begin{equation}\label{cardyn}
    \mathbf{X}_{v,k+1}=\mathbf{X}_{v,k}+\Delta T\mathbf{U}_{v,k}
\end{equation}
At any instant, we assume that the velocity of the vehicle in its body-fixed frame lies in the set 
%$\mathcal{B}_v=\{v\ :\ v\in\mathbb{R}^2,\ ||v||\leq\bar V,\ \arcsin(\frac{e^T_1v}{||v||})\in \underbar\Theta,\bar\Theta \}$
$\mathcal{B}_v=\{v\ :\ v\in\mathbb{R}^2,\ ||v||\leq\bar V,\ \arcsin(\frac{e^T_1v}{||v||})\in [\underline\Theta,\bar\Theta] \}$
The bounds $\bar V, \underline\Theta$ and $\bar\Theta$ are assumed to be known beforehand.
\begin{figure}%[h]
    \centering
    \includegraphics[scale=0.52]{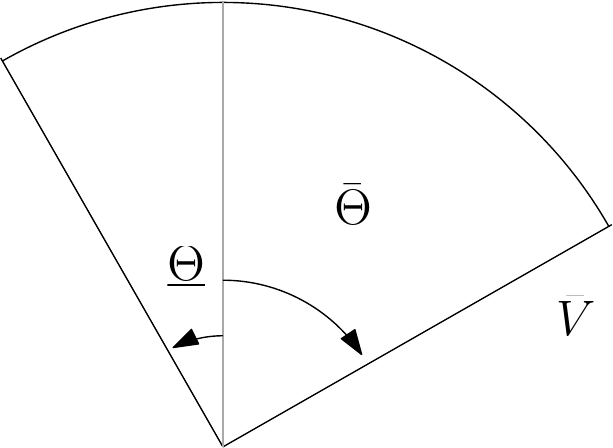}
    \caption{$\mathcal{B}_v$}
    \label{fig:car}
\end{figure}

%%%%%%%%%%%%%%%%%%%%%%%%%%%%%%%%%%%%%

\subsection{Receding Horizon Control Problem for Tracking MPC}\label{RHC}
With design parameters $N$ (prediction horizon), $Q$ (state cost matrix) and $R$ (input cost matrix), we wish to solve the following optimization problem $(P)$ at time $t$,\\\\
minimize $\sum_{k=0}^N ||\mathbf{X}_k-\mathbf{X}_{k}^{ref}||^2_{Q}+||\mathbf{U}_k||^2_R$\\
subject to
\begin{align*}
    &\mathbf{X}_{k+1}=A_T\mathbf{X}_k+B_T\mathbf{U}_k +G_T\\
&\mathbf{X}_k\in\mathcal{X},\ \mathbf{X}_N\in \mathcal{X}_f,\ \mathbf{U}_k\in\mathcal{U}\\
&\mathbf{X}_0=\mathbf{X}(t)
\end{align*}
$\mathcal{X}$ and $\mathcal{U}$ are the sets of feasible states and control inputs respectively. We assume that these sets are convex and preferably, polyhedral. $\mathcal{X}_f$ is the  terminal constraint set and will be constructed in the sequel, along with a heuristic for designing the reference trajectory $\mathbf{X}^{ref}$ to make the quadcopter chase the ground vehicle.\\\\
The MPC algorithm prescribes using $U^*_0$ yielded by solving $(P)$ to obtain the next state $\mathbf{X}(t+\Delta T)$ using dynamics (\ref{qmodel}), and then solving $(P)$ again with $X_0=\mathbf{X}(t+\Delta T)$.
%%%%%%%%%%%%%%%%%%%%%%%%%%%%%

\subsection*{Construction of Terminal Set $\mathcal{X}_f$}
Let $X_c$ be the current state of the ground vehicle. Define the set
\begin{equation}\label{bf}
   \mathcal{B}_f=X_c\oplus\{x\in\mathbb{R}^2:||x||\leq \bar{V}N\Delta T \} 
\end{equation}
\begin{lemma}
For a vehicle governed by dynamics (\ref{cardyn}) and for the set $\mathcal{B}_f$ constructed as in (\ref{bf}), the following are true:
\begin{enumerate}
    \item Set $\mathcal{B}_f$ is convex.
    \item The vehicle positions lie within this set for all instants over the prediction horizon $N$.
\end{enumerate}
\end{lemma}
\textbf{\textit{Proof}}\\
Part 1)\\
$\mathcal{B}_{f}$ is a circle centered at $X_c$, which is convex.\\\\
Part 2)\\
At any instant $k$,
\small
\begin{align*}
\mathbf{X}_{vk}-\mathbf{X}_{vk-1}&=\Delta T\mathbf{U}_{vk-1}=\Delta T\begin{bmatrix}
||v_{k-1}||\cos(\phi_{k-1}-\delta_{k-1})\\
||v_{k-1}||\sin(\phi_{k-1}-\delta_{k-1})
\end{bmatrix}\\
\Rightarrow||\mathbf{X}_{vk}&-\mathbf{X}_{vk-1}||\leq\bar{V}\Delta T
\end{align*}
\normalsize
Using a telescopic sum and the triangle inequality, we get 
\small
\begin{align*}
||X_{vk}-X_{c}||&\leq k\Delta T \bar{V}\leq N\Delta T\bar{V}\ \ \forall k=1,2,..N\\
\Rightarrow X_{vk}&\in\mathcal{B}_f\quad \forall k=1,2,..N
\end{align*}
\normalsize
$\blacksquare$\\\\
Construct the translational part of the terminal constraint set as
$$\mathcal{X}_{ft}=\mathcal{B}_f\times[0,H]\times [e_2\  e_6\ e_{10}]^T\circ\mathcal{X}$$
where the $e_{i}$s are among the standard basis vectors of $\mathbb{R}^{10}$ and $H$ is the height from the ground within which we consider the quadcopter to have "captured" the vehicle.\\
Now define following matrix which plucks out the rotational states from state vector $X$

\scriptsize
$$\Phi=\begin{bmatrix}
0 & 0 & 0 & 0 & 0 & 0 & 0 & 0 & 0 & 0\\
0 & 0 & 0 & 0 & 0 & 0 & 0 & 0 & 0 & 0\\
0 & 0 & 1 & 0 & 0 & 0 & 0 & 0 & 0 & 0\\
0 & 0 &  0& 1 & 0 & 0 & 0 & 0 & 0 & 0\\
0 & 0 & 0 & 0 & 0 & 0 & 0 & 0 & 0 & 0\\
0 & 0 & 0 & 0 & 0 & 0 & 0 & 0 & 0 & 0\\
0 & 0 & 0 & 0 & 0 & 0 & 1 & 0 & 0 & 0\\
0 & 0 & 0 & 0 & 0 & 0 & 0 & 1 &0 & 0\\
0 & 0 & 0 & 0 & 0 & 0 & 0 & 0 & 0 & 0\\
0 & 0 & 0 & 0 & 0 & 0 & 0 & 0 & 0 & 0\\
\end{bmatrix}
$$
\normalsize
We use this to define the rotational part of the terminal constraint set
$$\mathcal{X}_{fr}=\Phi\circ\mathcal{X}$$ and finally, use cartesian product to construct the terminal constraint set
\begin{equation}\label{tset}
\mathcal{X}_f=(\mathcal{X}_{ft}\times\mathcal{X}_{fr})\cap \mathcal{X}
\end{equation}\\

\begin{lemma}
The terminal constraint set given by (\ref{tset}) is control invariant for the quadcopter dynamics described by (\ref{qmodel}) with the terminal controller
\small
\begin{align}\label{uf}
    U_f(\mathbf{X}_k)&=\textrm{sup}\{\begin{bmatrix}
    u\\0
    \end{bmatrix}: \begin{bmatrix}
    u\\0
    \end{bmatrix}\in\mathcal{U},\nonumber\\
    &\frac{[e_1\ e_5]^T(A_T-I)\mathbf{X}_k+B^\dagger_Tu}{||[e_1\ e_5]^T(A_T-I)\mathbf{X}_k+B^\dagger_Tu||}=\frac{X_c-[e_1\ e_5]^T\mathbf{X}_k}{||X_c-[e_1\ e_5]^T\mathbf{X}_k||} \}
\end{align}
\normalsize
where $B^\dagger_T=[e_1\ e_5]^TB_T[e_1\ e_2]$, if the following conditions are met
\begin{enumerate}
    \item $[-\bar{V}\Delta T, \bar{V}\Delta T]\subseteq\Delta x\cap\Delta y$ where 
\small
$$\Delta x=\ e_1^T\circ((A_T-I)\circ\mathcal{X}\oplus B_T\circ\mathcal{U})$$
$$\Delta y=\ e_5^T\circ((A_T-I)\circ\mathcal{X}\oplus B_T\circ\mathcal{U})$$

\normalsize
\item The quadcopter platform has large enough $\mathcal{X}$ such that $$\exists U\in U_f(\mathbf{X}_k): A_T\mathbf{X}_k+B_TU +G_T\in\mathcal{X}$$
\item $\mathcal{U}$ contains an open ball around the origin $\mathbf{0}$
\end{enumerate}

\end{lemma}
\textbf{\textit{Proof}}\\
Suppose that the state of the quadcopter at the $k$th instant belongs to the terminal set at the $k$th instant, i.e., $\mathbf{X}_k\in\mathcal{X}_f$. We want to show that $\mathbf{X}_{k+1}\in\mathcal{X}_f$ on applying control (\ref{uf}).\\\\
First, observe that the subsystem of the quadcopter comprising the $x,\ y$ translational dynamics and the pitch, roll dynamics is controllable with inputs $\theta_{cmd},\phi_{cmd}$. This implies that $B^\dagger_T$ is invertible, and moreover with condition (3), this ensures that the set $U_f(\mathbf{X}_k)\neq \emptyset$. On applying this input,\ the quadcopter moves in the direction $\frac{X_c-[e_1\ e_5]^T\mathbf{X}_k}{||X_c-[e_1\ e_5]^T\mathbf{X}_k||}$.\\\\
At instant $k+1$, the horizontal distance between the quadcopter and the vehicle's new position $X'_c$ is given by
\small
\begin{align*}
||X'_c-[e_1\ e_5]^T\mathbf{X}_{k+1}||&=||X'_c-X_c+X_c-[e_1\ e_5]^T\mathbf{X}_{k+1}||\\
&\leq ||X'_c-X_c||+||(X_c-[e_1\ e_5]^T\mathbf{X}_{k})+\\&([e_1\ e_5]^T\mathbf{X}_{k}-[e_1\ e_5]^T\mathbf{X}_{k+1})||
\end{align*}
\normalsize
Condition (1) ensures that quadcopter's displacement in the direction $\frac{X_c-[e_1\ e_5]^T\mathbf{X}_k}{||X_c-[e_1\ e_5]^T\mathbf{X}_k||}$ exceeds $\bar{V}\Delta T$ and thus,
\small
$$||(X_c-[e_1\ e_5]^T\mathbf{X}_{k})+[e_1\ e_5]^T(\mathbf{X}_{k}-\mathbf{X}_{k+1})||\leq (N-1)\bar{V}\Delta T$$
\begin{align*}
\therefore\ ||X'_c-[e_1\ e_5]^T\mathbf{X}_{k+1}||&\leq ||X'_c-X_c||+(N-1)\bar{V}\Delta T\\&\leq N\bar{V}\Delta T\quad\quad\textrm{---(A)}
\end{align*}
\normalsize
Note that the altitude dynamics is decoupled from the other state dynamics and $\mathbf{X}_k\in\mathcal{X}_f$. So, applying $0$ thrust as prescribed by $U_f(\mathbf{X}_k)$ results in
\small$$e_9^T\mathbf{X}_{k+1}\leq H\quad\quad\textrm{---(B)}$$\normalsize
(A), (B) and condition (2) together give
$$\mathbf{X}_{k+1}\in\mathcal{X}_f$$
$\blacksquare$
\subsection*{Construction of Set of Feasible Initial States $\mathcal{X}_0$}
We construct $\mathcal{X}_0$ as the $N$-step reachable set to $\mathcal{X}_f$. For ease of computations, we construct the polyhedron that circumscribes the set $\mathcal{X}_f$ to get a (consequently) conservative estimate  $\Tilde{\mathcal{X}}_0\subset\mathcal{X}_0$.

\begin{theorem}
The MPC described in section (\ref{RHC}) is recursively feasible if $X(0)\in\mathcal{X}_0$ and the conditions of lemma 2 hold.
\end{theorem}
\textbf{\textit{Proof}}\\
Since $X(0)\in\mathcal{X}_0$, the RHC optimization problem $(P)$ is feasible over the first prediction horizon. Now suppose that $X(k\Delta T)\in\mathcal{X}_0$ and let $\{u^*_0,u^*_1,..,u^*_N\}$ be the optimal sequence obtained by solving $(P)$. Using the terminal controller $U_f(\mathbf{X}^*_N)$ from lemma 2, we construct the sequence $\{u^*_1, u^*_2,..,U_f(\mathbf{X}^*_N)\}$ which clearly yields a feasible solution to $(P)$ with $X_0=\mathbf{X}^*_{1}=X((k+1)\Delta T)$. Using induction on $k$, this shows that the MPC strategy is recursively feasible.\\
$\blacksquare$
\subsection{Reference Trajectory Generation for Tracking MPC}
For constructing the reference trajectory $X^{ref}$ that is passed to the MPC algorithm, we fit a minimum jerk trajectory between the quadcopter's current state and a point $H$ above an estimate of the vehicle's future position. The roll, pitch and yaw angles are set to $0$ at the terminal point. We now present a heuristic to get an estimate the vehicle's future position.
\subsection*{Updating $\mathcal{B}_v$}
We measure and store  time series data for $X_{vk}=[x_{vk}\ y_{vk}]^T$ and $U_{vk}=[v_{xk}\ v_{yk}]^T $ over time instants in  $[t-L\Delta T, t]$  where $t$ is the current time and $L$ is a design parameter. The velocity of the vehicle in the body-fixed frame is obtained as follows
\begin{align}
    v_k=\begin{bmatrix}
    \sqrt{v_{xk}^2+v_{yk}^2}\sin(\delta_k)\\
    \sqrt{v_{xk}^2+v_{yk}^2}\cos(\delta_k)
    \end{bmatrix}
\end{align}
where
\begin{align*}
    \delta_k&=\arcsin{\frac{v_{xk}}{\sqrt{v_{xk}^2+v_{yk}^2}}}-\phi_k\\
\end{align*}
and $\phi_k$ is the measured heading of the car.
\begin{figure}[h]
    \centering
    \includegraphics[scale=0.43]{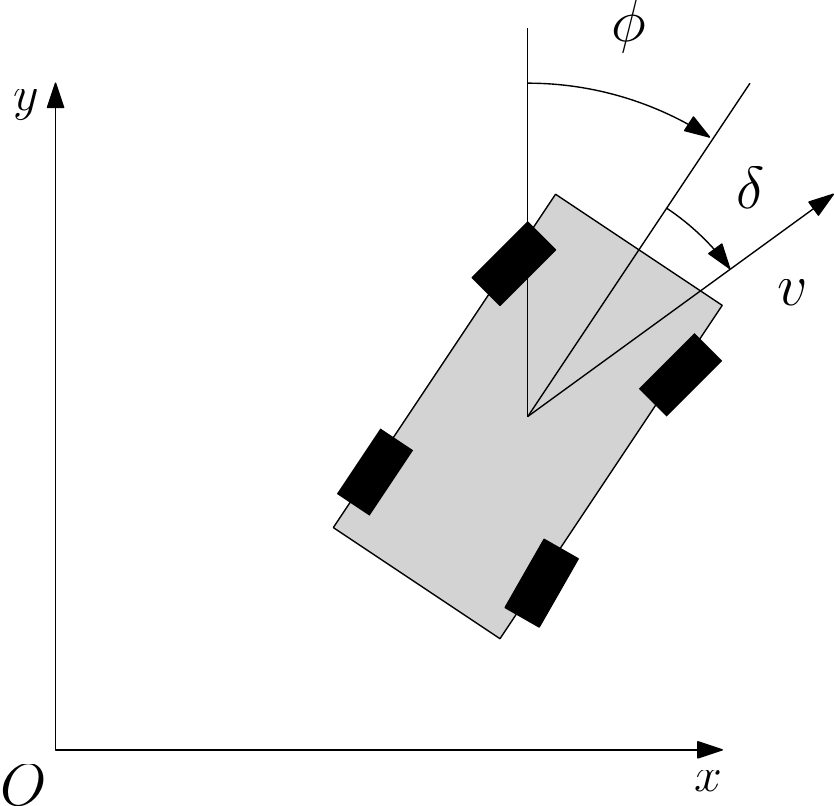}
    \caption{Heading angle and velocity direction of car}
    \label{fig:car}
\end{figure}
\\ Compute the average of the time-series data as $\Tilde{v}$,$\Tilde{\delta}$ and update the bounds $\bar{v}$, $\underline{\delta}$ and $\bar{\delta}$ as follows
\begin{align}
  \bar{v}&=\bar{V}(1-\beta_v)+\beta_v\Tilde{v}\\ 
 \underline{\delta}&=\underline{\Theta}(1-\underline{\beta})+\underline{\beta}\Tilde{\delta}\\
  \bar{\delta}&=\bar{\Theta}(1-\bar{\beta})+\bar{\beta}\Tilde{\delta}
\end{align}
$\bar{V},\underline{\Theta}$ and $\bar{\Theta}$ are bounds that are known beforehand while $\beta_v,\underline{\beta}$ and $\bar{\beta}$ are tuning parameters.
\subsection*{Prediction of the Vehicle's Position}

\begin{figure}[h]
    \centering
    \includegraphics[scale=0.38]{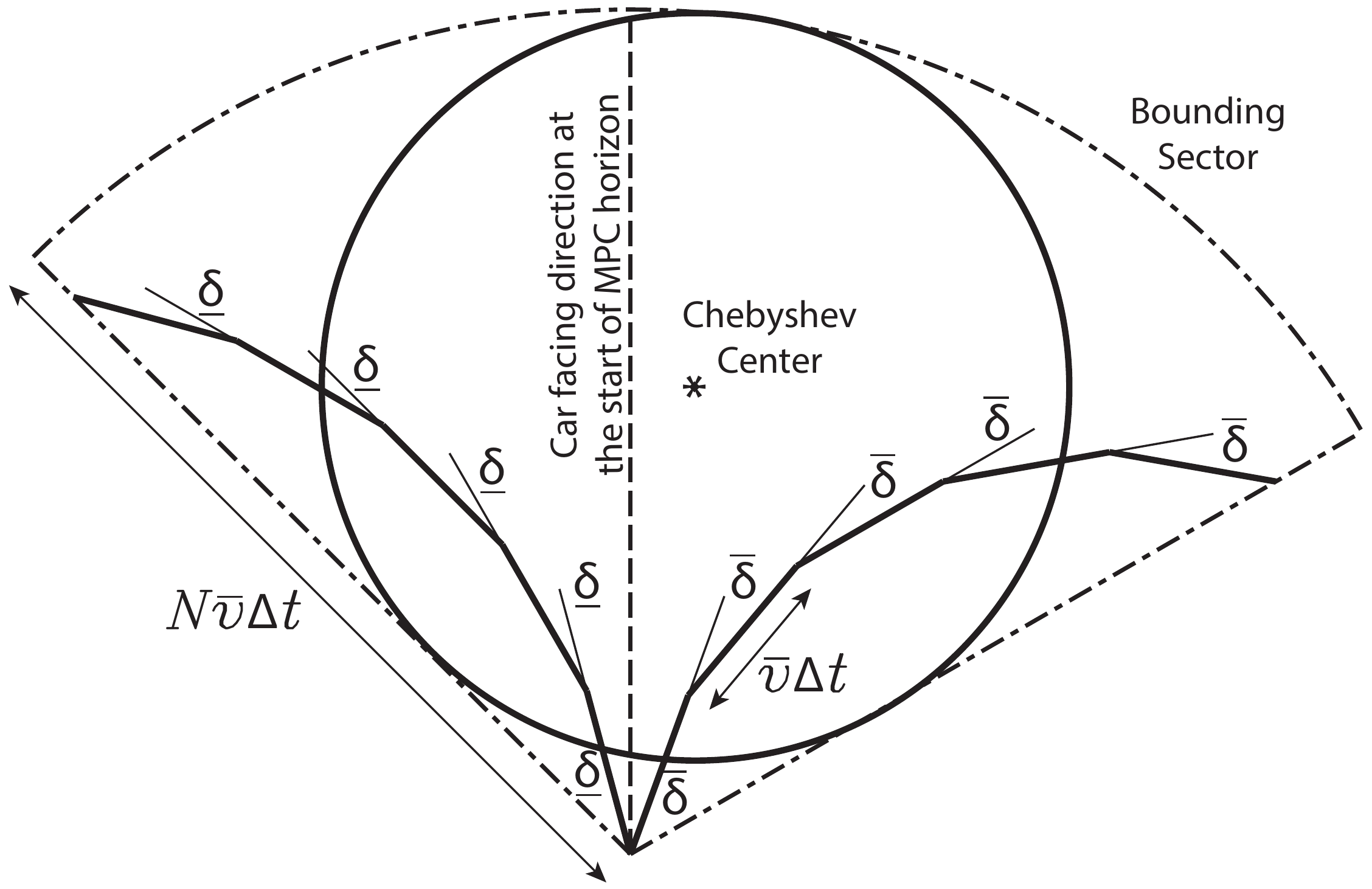}
    \caption{Prediction of car's position}
    \label{fig:prediction}
\end{figure}

Obtain a sequence of car states $\{\underline{X}_{vk}\}_{k=0}^{N}$ using dynamics (\ref{cardyn}) initialized with the current measured state of the car, over the prediction horizon $N$, using inputs $\underline{U}_k$ given by

\small
$$
\underline{U}_k=\begin{bmatrix}
\bar{v}\cos(\phi_k-\underline{\delta})\\
\bar{v}\sin(\phi_k-\underline{\delta})
\end{bmatrix}
$$
\normalsize\\
Similarly, obtain a sequence of car states $\{\overline{X}_{vk}\}_{k=0}^{N}$ using dynamics (\ref{cardyn}) initialized with the current measured position of the car, over the prediction horizon $N$, using inputs $\overline{U}_k$ given by
\small
$$
\overline{U}_k=\begin{bmatrix}
\bar{v}\cos(\phi_k-\bar{\delta})\\
\bar{v}\sin(\phi_k-\bar{\delta})
\end{bmatrix}
$$
\normalsize\\
Let $\underline{X}_{v0}=\overline{X}_{v0}=X_c$ and compute
\small
$$\underline{\theta}=\arcsin{\left(\frac{e_1^T(\underline{X}_{vN+1}-X_0)}{||\underline{X}_{vN+1}-X_0||}\right)}$$
$$\bar{\theta}=\arcsin{\left(\frac{(\overline{X}_{vN+1}-X_0)^Te_1}{||\overline{X}_{vN+1}-X_0||}\right)}
$$
\normalsize
Define the set
\begin{align}\label{cf}
    \mathcal{C}_f=\mathcal{C}_{f1}\cup\mathcal{C}_{f2}
\end{align}
where
\small
\begin{align*}
    \mathcal{C}_{f1}&=X_0\oplus\{x\in\mathbb{R}^2 :x=r\begin{bmatrix}
   \sin\theta\\ \cos\theta \end{bmatrix}, r\in[0,\bar{v}N\Delta T], \theta\in[\underline{\theta},\bar{\theta}]\}\\
   \mathcal{C}_{f2}&=X_0\oplus conv(0,\bar{v}N\Delta T\begin{bmatrix}
   \cos{\underline{\theta}}\\ \sin{\underline{\theta}} \end{bmatrix},\bar{v}N\Delta T\begin{bmatrix}
   \cos{\bar{\theta}}\\ \sin{\bar{\theta}} \end{bmatrix})
\end{align*}
\normalsize\\
\begin{lemma}
 The set $\mathcal{C}_f$ constructed as in (\ref{cf}), is bounded and convex
\end{lemma}
\textbf{\textit{Proof}}\\
Observe that by construction, $\mathcal{C}_f\subset\mathcal{B}_f$ and is hence, bounded.\\
$\mathcal{C}_{f1}$ is a sector of a circle that subtends an angle $\bar{\theta}-\underline{\theta}$ at the centre $X_c$ whereas $\mathcal{C}_{f2}$ is a triangle with the same vertices as that of the sector $\mathcal{C}_{f1}$.\\\\
Case 1: $\bar{\theta}-\underline{\theta}\leq\pi$\\
$\mathcal{C}_{f1}$ is convex and $\mathcal{C}_{f2}\subset \mathcal{C}_{f1}$.\\ Thus $\mathcal{C}_f=\mathcal{C}_{f1}\cup\mathcal{C}_{f2}=\mathcal{C}_{f1}$\\
$\Rightarrow \mathcal{C}_f$ is convex\\\\
Case 2: $\bar{\theta}-\underline{\theta}>\pi$\\
In this case, $\mathcal{C}_{f2}\not\subset\mathcal{C}_{f1}$ and the set $\mathcal{C}_f=\mathcal{C}_{f1}\cup\mathcal{C}_{f2}$ forms a segment of the circle centered at $X_c$, which is convex.\\
$\Rightarrow \mathcal{C}_f$ is convex\\\\
Cases 1 and 2 are exhaustive and hence, $\mathcal{C}_f$ is indeed convex.\\
$\blacksquare$\\
\begin{prop}
The vehicle's future position is estimated as the Chebyshev center of $\mathcal{C}_f$ (Figure \ref{fig:prediction}) and this estimate belongs to $\mathcal{C}_f\subset\mathcal{B}_f$.
\end{prop}
\textbf{\textit{Proof}}\\
Lemma 3 directly gives the desired result.\\
$\blacksquare$

%%%%%%%%%%%%%%%%%%%%%%%%%%
\section{Simulations}

Two simulated chasing experiments were carried out with a ROS-based quadcopter Simulation tool. The tool takes control commands via ROS communication, simulates the flight via propagating through a high fidelity quad-copter vehicle dynamics model, and returns full vehicle states. For the first simulated test, quad-copter was ordered to chase a ground vehicle moving in a circular trajectory. For the second simulated experiment, the quadcopter chased a car that randomly drove in a bounded square virtual test field.

%%%%%%%%%%%%%%%%%%%%%%%%%%%%%%
\subsection{Simulation-1 Result}

\begin{figure}[H]
    \centering
    \includegraphics[scale=0.5]{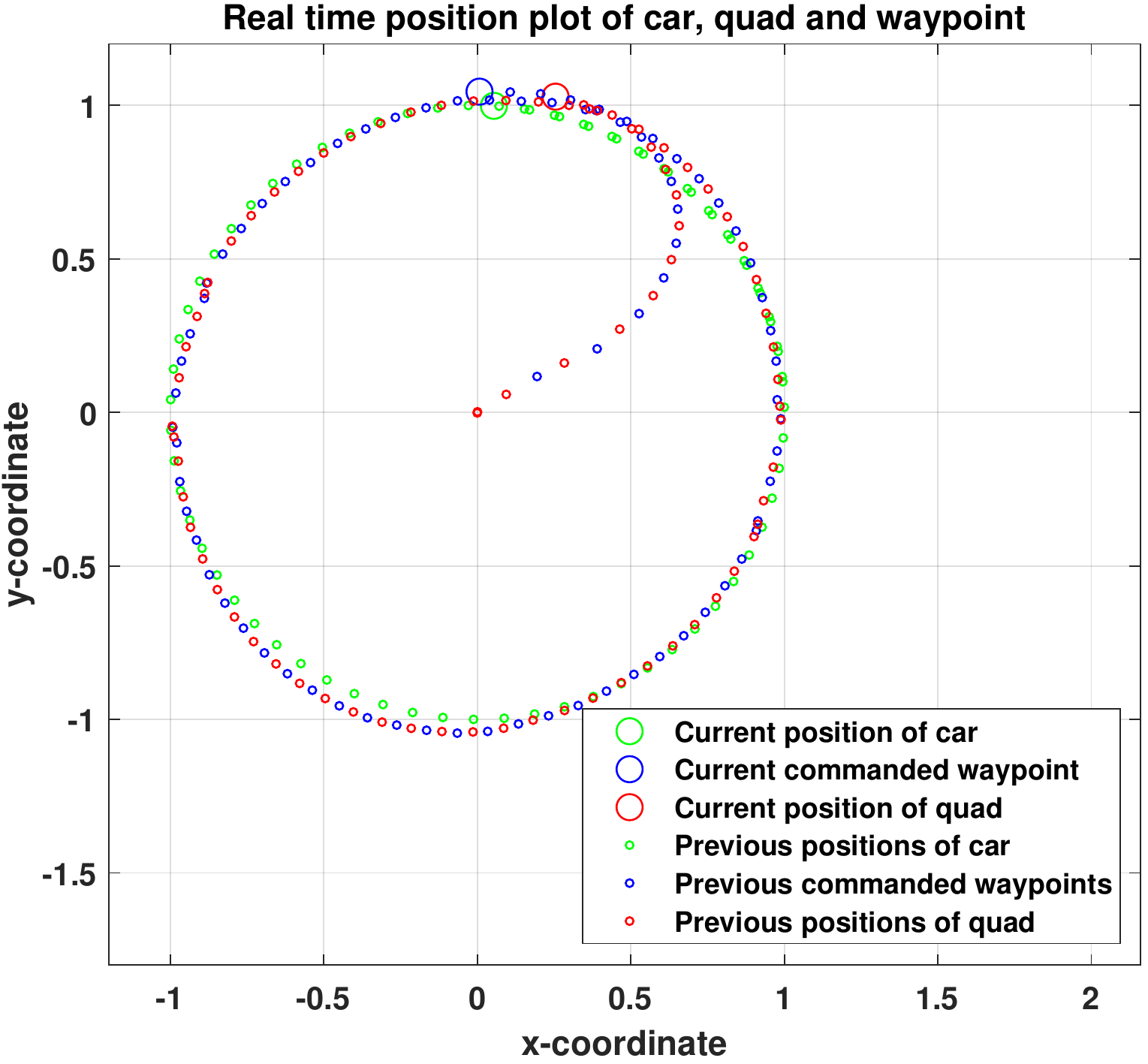}
    \caption{Tracking a ground vehicle running in circular trajectory}
    \label{fig:simulation1trajectory}
\end{figure}

\begin{figure}[H]
    \centering
    \includegraphics[scale=0.4]{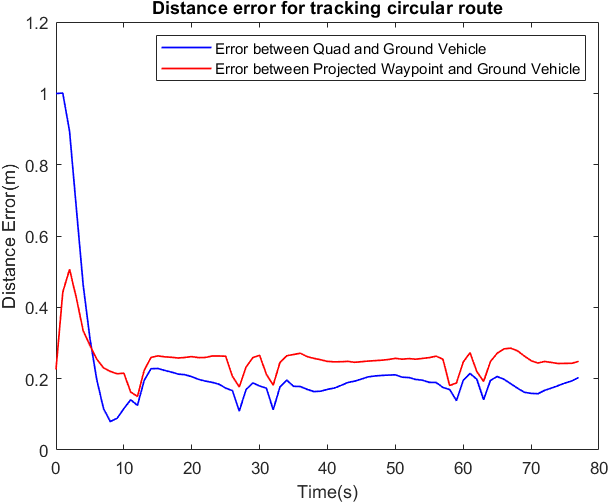}
    \caption{Tracking distance error for simulation 1}
    \label{fig:simulation1trackingerror}
\end{figure}

Figure \ref{fig:simulation1trajectory} shows the trajectory of a simulated quadcopter chasing a ground vehicle simulated to run in a circular path. Figure \ref{fig:simulation1trackingerror} shows a plot for the tracking distance error against time. We observe that the distance error quickly converges after MPC controller starts working and that we manage to contain the steady state error within about 0.25 m.

% \begin{figure}

% \end{figure}

%%%%%%%%%%%%%%%%%%%%%%%%%%%%%%%
\subsection{Simulation-2 Result}
%%% @KJ Add the  random work graphs here if you find one.

\begin{figure}[H]
    \centering
    \includegraphics[width=\columnwidth]{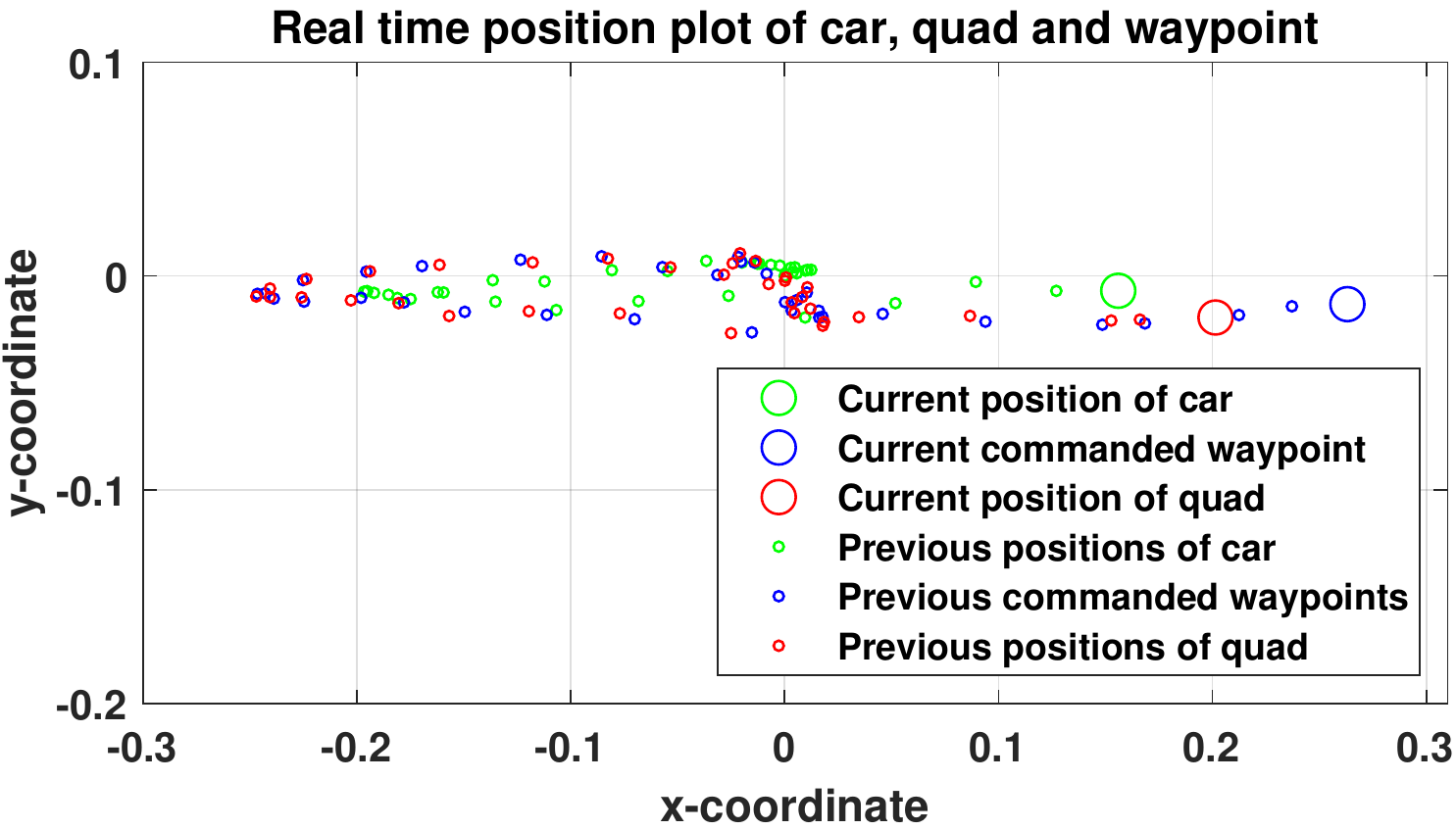}
    \caption{Tracking a vehicle with random acceleration and steering inputs}
    \label{fig:simulation2trajectory}
\end{figure}

\begin{figure}[H]
    \centering
    \includegraphics[scale=0.4]{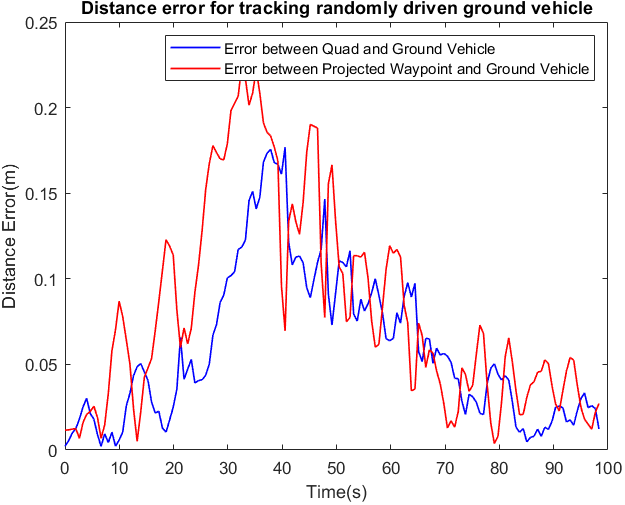}
    \caption{Tracking distance error for simulation 2}
    \label{fig:simulation2trackingerror}
\end{figure}

Figure \ref{fig:simulation2trajectory} shows the trajectory of a simulated quadcopter chasing a ground vehicle simulated to randomly drive in a bounded square field. Figure \ref{fig:simulation2trackingerror} shows a plot for the tracking distance error against time. From the graph, we observe that the tracking error would increase when the vehicle make sharp turns, but the MPC controller manages to keep that value under 0.25 m.

%%%%%%%%%%%%%%%%%%%%%%%%%%%%%%

\section{Experiments}
\subsection{Experiment Setup}
\begin{figure}[H]
    \centering
    \includegraphics[scale=0.05]{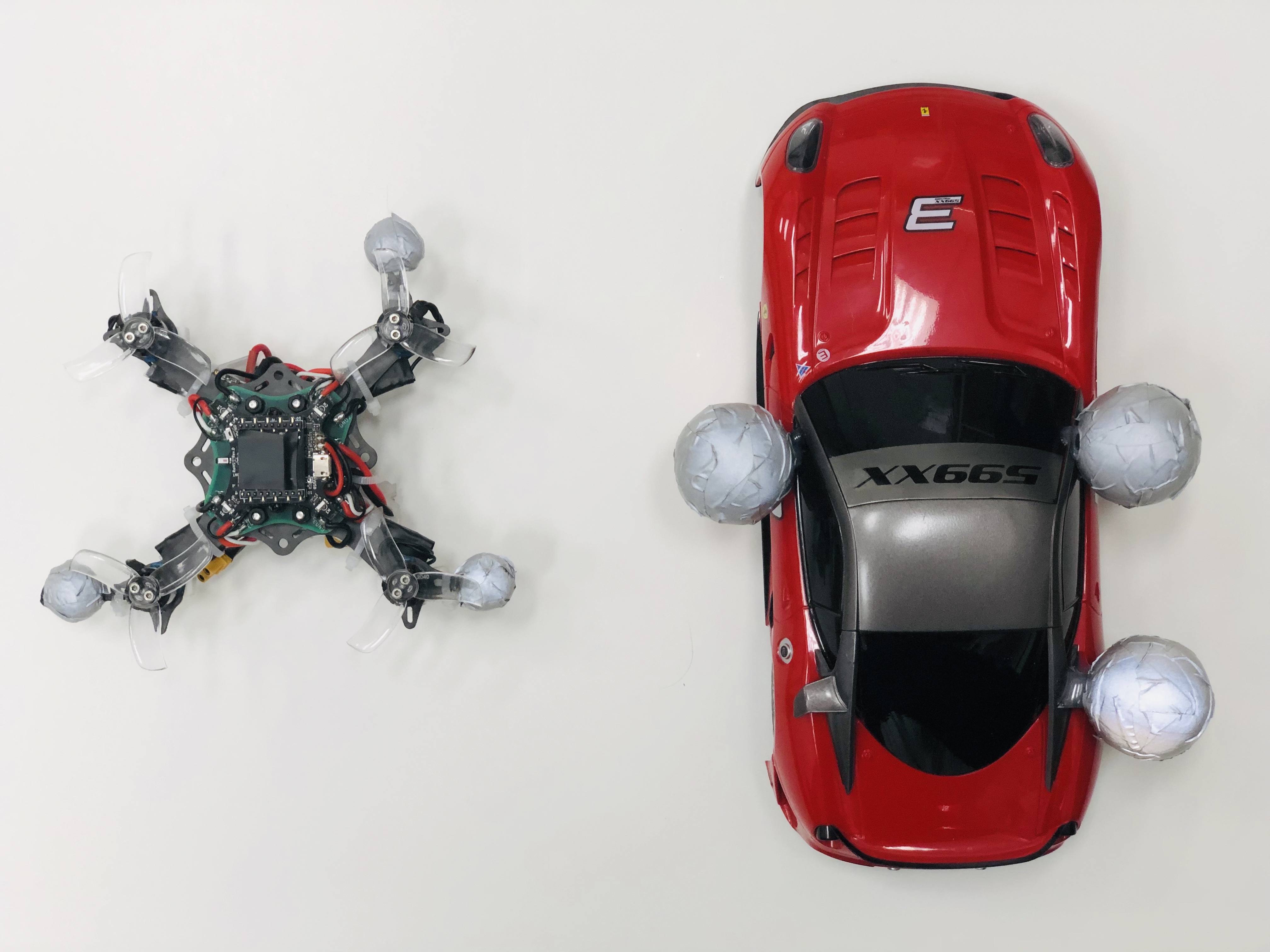}
    \caption{Vehicles used for the experiment}
    \label{fig:vehicle}
\end{figure}

\begin{figure}[H]
    \centering
    \includegraphics[width=\columnwidth]{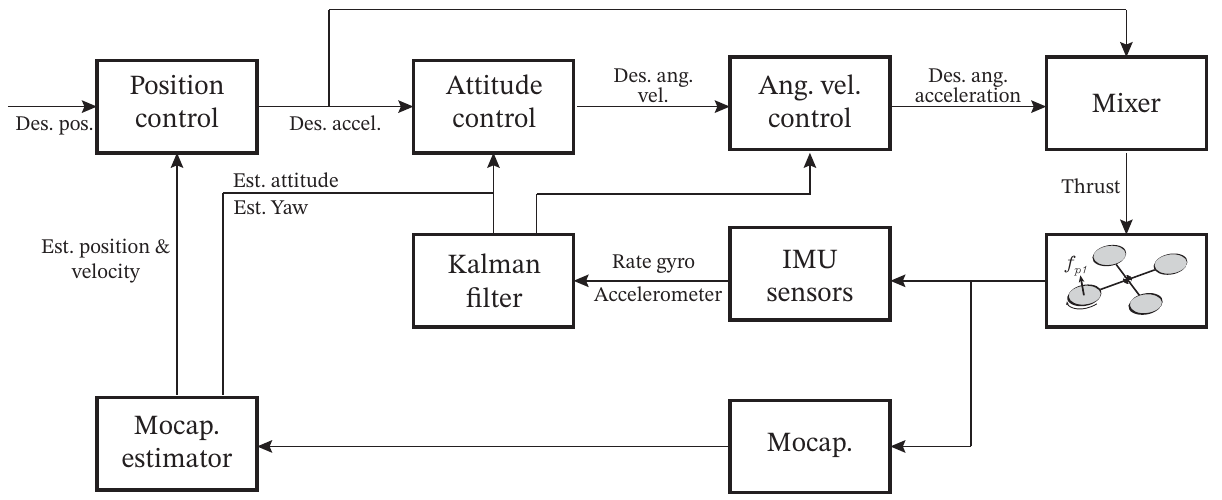}
    \caption{Low-level Control Architecture}
    \label{fig:llc}
\end{figure}

In this experiment, a  RC car is manually controlled by a human driver and a quadcopter flies autonomously to chase the RC car. The motion of the car and the quad-copter is collected by the motion capture system and sent to a real-time MPC program, which generates a trajectory in terms of desired way-points and corresponding way-point velocities. The trajectory is then interpreted as radio commands and sent to the quadcopter to close the control loop. We work with a crazyflie-based quadcopter platform, which is equipped with a robust, low-level trajectory tracking controller. Figure \ref{fig:llc} depicts the low-level control architecture of the miniquad platform.

\begin{figure}%[h]
    \centering
    \includegraphics[width=7cm]{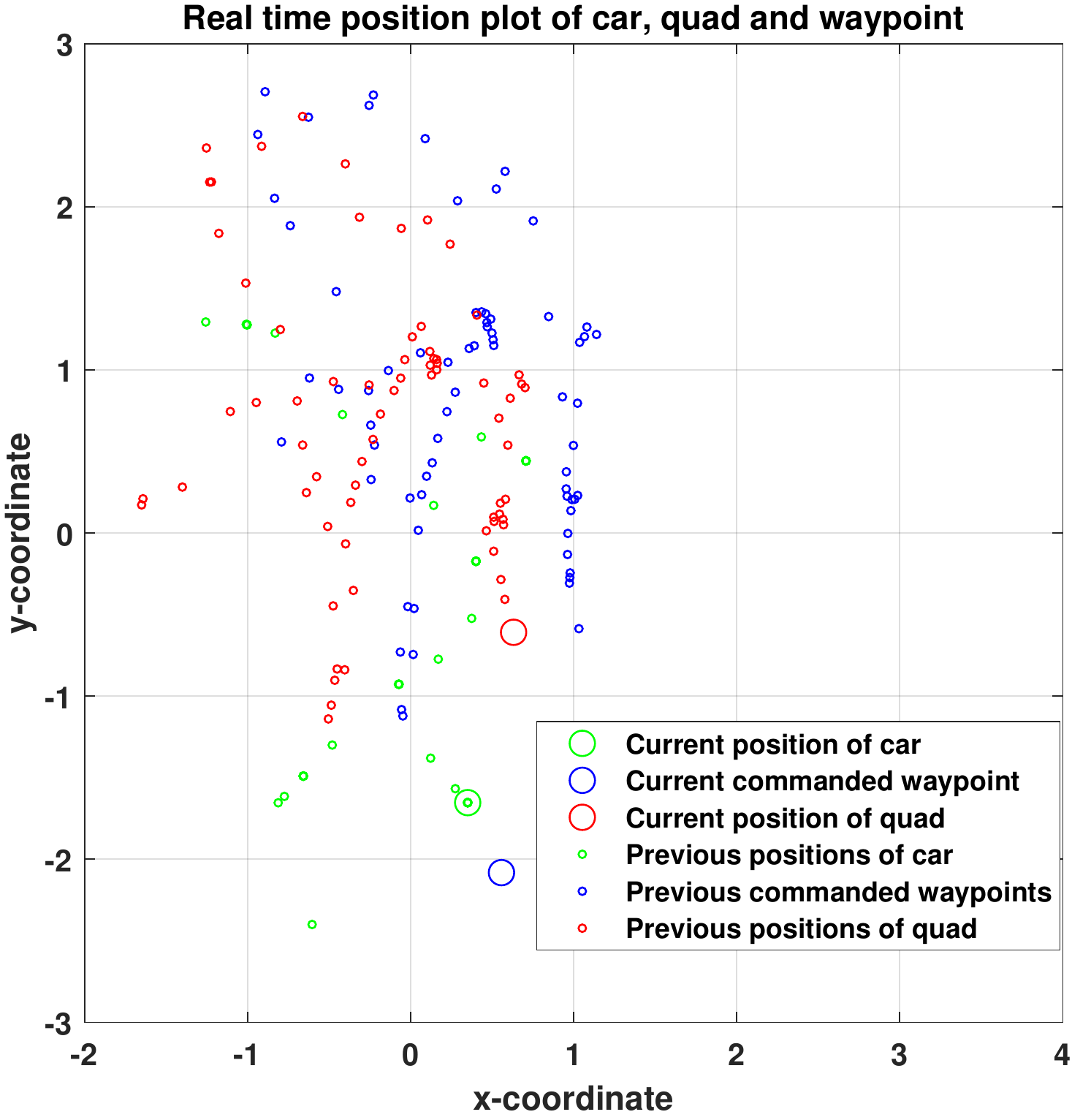} % Just use 
    \caption{Experimental tracking of a remote controlled car controlled by a human operator. The car is being chased by a quadcopter following waypoints generated by our predictive control algorithm}
    \label{fig:experiment1trajectory}
\end{figure}
\begin{figure}%[h]
    \centering
    \includegraphics[scale=0.5]{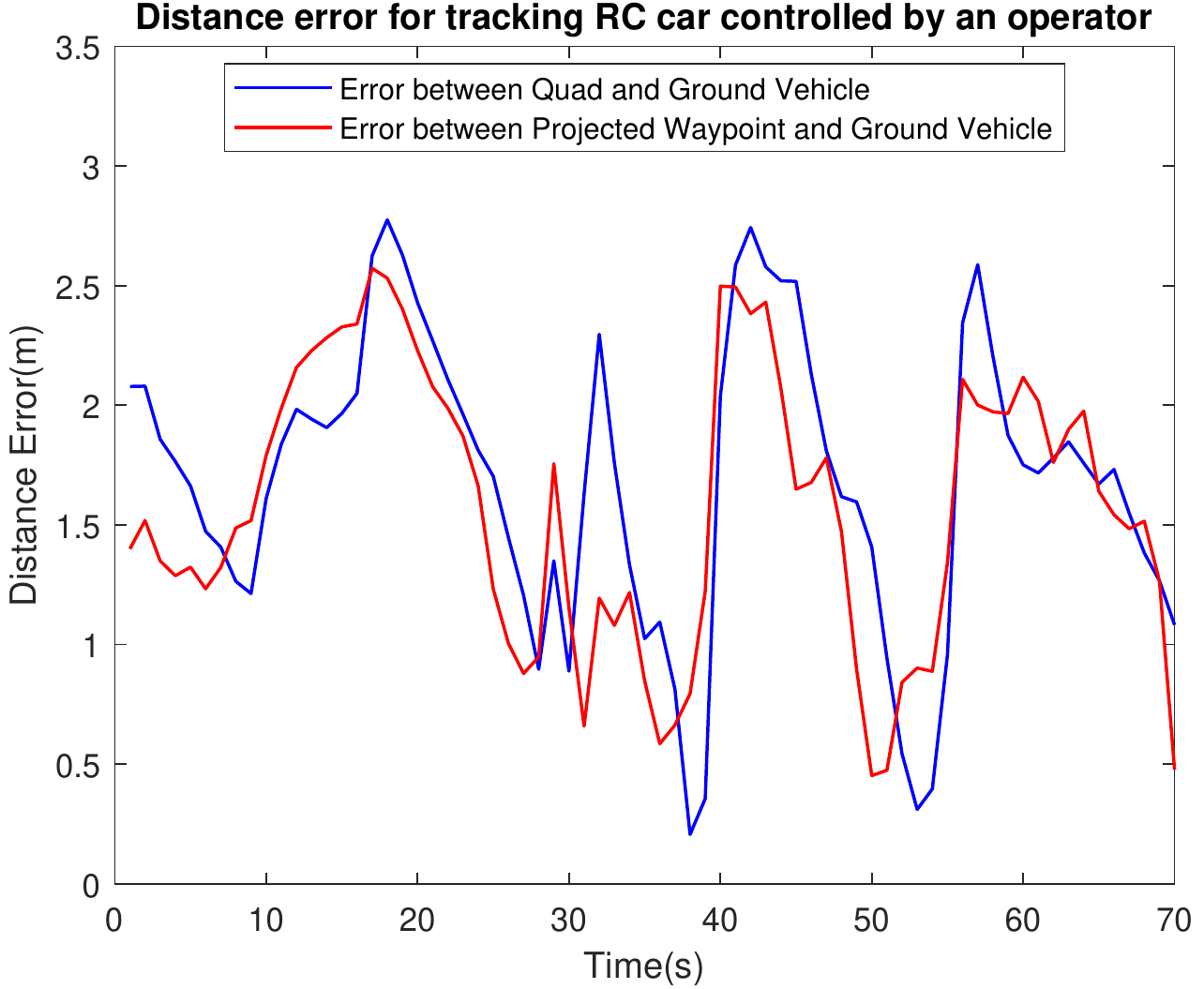}
    \caption{Tracking distance error for the experiment}
    \label{fig:experiment1trackingerror}
\end{figure}

\subsection{Experiment Result}

Figure \ref{fig:experiment1trajectory} shows the trajectory of an actual quadcopter chasing a remotely controlled ground vehicle that is driven by an operator in our laboratory space. Figure \ref{fig:experiment1trackingerror} shows a plot for the tracking distance error against time. From the graph, we observe that the tracking error increases when the vehicle accelerates, resulting in peaks. As time progresses, the predictive controller manages to reduce the error to under 1.0 m.

%%%%%%%%%%%%%%%%%%%%%%%%%%%%%%

The video of the experiment can be seen here: \small\url{https://goo.gl/eUEq2W}\normalsize

We can clearly observe the difference in performance between experiments and simulations. The experimental tracking error is about an order of magnitude higher than that of simulations. Following are the reasons for the inadequate performance of the experimental chase:
\begin{enumerate}
    \item Model mismatch due to linearization of dynamics of the quadcopter in our control algorithm.
    \item Time delays in sending and receiving radio messages to and from the quadcopter during experiments.
    \item Measurement noise in sensors which leads to imperfect waypoint tracking.
\end{enumerate}

% \addtolength{\textheight}{-12cm}

\section{Conclusions and Future Work}
In this paper, we focused on designing a predictive controller for an aerial vehicle to chase a ground vehicle. We assume a model for the ground vehicle for estimating its position in future time instants. These estimates are used to generate waypoints for our aerial vehicle so that it can track the ground vehicle with an objective of minimizing the tracking error in distance.

We proved feasibility of the optimization problem that we solve. We then implemented this control algorithm for simulations and experiments.

In simulations, we achieved good tracking performance with tracking errors smaller than 0.25 m. However, in the case of experiments, we had much larger tracking errors owing to non-idealities such as model mismatch, time delays, and sensor noise.

In future work, we will characterize the tracking error bound, and incorporate probabilistic beliefs into our prediction strategy. Furthermore, while we have incorporated slacks variables in our optimization problems for this work, we could also tackle model mismatch via a robust MPC strategy. Lastly, waypoint tracking will be improved by commanding velocities and accelerations in addition to positions.

\section*{Acknowledgement}
The experimental test bed at the HiPeRLab is the result of contributions of many people, a full list of which can be found at \url{hiperlab.berkeley.edu/members/}. 

\bibliographystyle{IEEEtran}
\bibliography{references}

\begin{thebibliography}{1}
\providecommand{\url}[1]{#1}
\csname url@samestyle\endcsname
\providecommand{\newblock}{\relax}
\providecommand{\bibinfo}[2]{#2}
\providecommand{\BIBentrySTDinterwordspacing}{\spaceskip=0pt\relax}
\providecommand{\BIBentryALTinterwordstretchfactor}{4}
\providecommand{\BIBentryALTinterwordspacing}{\spaceskip=\fontdimen2\font plus
\BIBentryALTinterwordstretchfactor\fontdimen3\font minus
  \fontdimen4\font\relax}
\providecommand{\BIBforeignlanguage}[2]{{%
\expandafter\ifx\csname l@#1\endcsname\relax
\typeout{** WARNING: IEEEtran.bst: No hyphenation pattern has been}%
\typeout{** loaded for the language `#1'. Using the pattern for}%
\typeout{** the default language instead.}%
\else
\language=\csname l@#1\endcsname
\fi
#2}}
\providecommand{\BIBdecl}{\relax}
\BIBdecl

\bibitem{silva2018vtol}
C.~Silva, W.~R. Johnson, E.~Solis, M.~D. Patterson, and K.~R. Antcliff, ``Vtol
  urban air mobility concept vehicles for technology development,'' in
  \emph{2018 Aviation Technology, Integration, and Operations Conference},
  2018, p. 3847.

\bibitem{hoang2017vision}
T.~Hoang, E.~Bayasgalan, Z.~Wang, G.~Tsechpenakis, and D.~Panagou,
  ``Vision-based target tracking and autonomous landing of a quadrotor on a
  ground vehicle,'' in \emph{2017 American Control Conference (ACC)}.\hskip 1em
  plus 0.5em minus 0.4em\relax IEEE, 2017, pp. 5580--5585.

\bibitem{teuliere2011chasing}
C.~Teuliere, L.~Eck, and E.~Marchand, ``Chasing a moving target from a flying
  uav,'' in \emph{2011 IEEE/RSJ International Conference on Intelligent Robots
  and Systems}.\hskip 1em plus 0.5em minus 0.4em\relax IEEE, 2011, pp.
  4929--4934.

\bibitem{archive2019}
\BIBentryALTinterwordspacing
S.~Choudhury and M.~J. Kochenderfer, ``Dynamic real-time multimodal routing
  with hierarchical hybrid planning,'' \emph{CoRR}, vol. abs/1902.01560, 2019.
  [Online]. Available: \url{http://arxiv.org/abs/1902.01560}
\BIBentrySTDinterwordspacing

\bibitem{borrelli2017predictive}
F.~Borrelli, A.~Bemporad, and M.~Morari, \emph{Predictive control for linear
  and hybrid systems}.\hskip 1em plus 0.5em minus 0.4em\relax Cambridge
  University Press, 2017.

\bibitem{bouffard2012board}
P.~Bouffard, ``On-board model predictive control of a quadrotor helicopter:
  Design, implementation, and experiments,'' UC Berkeley, Department of EECS,
  Tech. Rep., 2012.

\bibitem{ccta}
J.~{Dentler}, S.~{Kannan}, M.~A.~O. {Mendez}, and H.~{Voos}, ``A real-time
  model predictive position control with collision avoidance for commercial
  low-cost quadrotors,'' in \emph{2016 IEEE Conference on Control Applications
  (CCA)}, Sep. 2016, pp. 519--525.

\end{thebibliography}

% \begin{thebibliography}{99}
% \bibitem{c3} Vidal, R., Rashid, S., Sharp, C., Shakernia, O., Kim, J. and Sastry, S., \textit{Pursuit-Evasion Games with Unmanned Ground and Aerial Vehicles}, 2001 IEEE International Conference on Robotics and Automation, 21-26 May 2001

% \bibitem{c4} Teuliere, C., Eck, L. and Marchand, E., \textit{Chasing a moving target from a flying UAV}, 2011 IEEE/RSJ International Conference on Intelligent Robots and Systems, 25-30 Sept. 2011

% \bibitem{c5} Hoang, T., Bayasgalan, E., Wang, Z., Tsechpenakis, G. and Panagou, D., \textit{Vision-based target tracking and autonomous landing of a quadcopter on a ground vehicle}, 2017 American Control Conference, 24-26 May 2017

% \bibitem{c1} Bouffard, P., \textit{On-board Model Predictive Control of a Quadcopter Helicopter: Design, Implementation, and Experiments}, Technical Report No. UCB/EECS-2012-241, Electrical Engineering and Computer Sciences, University of California at Berkeley, 2012
% \bibitem{c2}Dentler J., Kannan S.,  Mendez M.A.O., Voos H.,\textit{A real-time model predictive position control with collision avoidance for commercial low-cost quadcopters},  2016 IEEE Conference on Control Applications (CCA), 19-22 Sept. 2016

% \bibitem{c6} Borrelli, F., Bemporad, A. and Morari, M., \textit{Predictive Control for Linear and Hybrid Systems}, Cambridge University Press, 2017
% \end{thebibliography}

\balance
\end{document}